%% file: main.tex
\documentclass[10pt,twocolumn,letterpaper]{article}

\usepackage{cvpr}
\usepackage{flushend}
\usepackage{graphicx}
\usepackage{times}
\usepackage{lipsum}
\usepackage{microtype}
\usepackage{tikz}  
\usepackage{epsfig}
\usepackage{graphicx}
\usepackage{amsmath}
\usepackage{amssymb}
\usepackage{float}
\usepackage{booktabs} 
\usepackage{array}    
\usepackage{colortbl} 
\usepackage{color, xcolor} 
\usepackage[pagebackref=true,breaklinks=true,colorlinks,bookmarks=false]{hyperref}
\usepackage{enumitem}
\setlist[itemize]{noitemsep, topsep=0.5pt, partopsep=0.5pt}
\setlist[enumerate]{noitemsep, topsep=0.5pt, partopsep=0.5pt}
\usepackage{fancyhdr}
\begin{document}
\title{ChatGPT Meets Iris Biometrics}

\author{Parisa Farmanifard and Arun Ross\\
Michigan State University, USA\\
{\tt\small \{farmanif,rossarun\}@msu.edu}
}

\maketitle
\begin{tikzpicture}[remember picture, overlay]
\node[anchor=north, text width=\paperwidth, align=center] at (current page.north) [shift={(0,-1.5cm)}] {\textcolor{red}{P. Farmanifard and A. Ross, “ChatGPT Meets Iris Biometrics,” \\Proc. of International Joint Conference on Biometrics (IJCB), (Buffalo, USA), September 2024.}
};
\end{tikzpicture}
\thispagestyle{fancy} 

\begin{abstract}
This study utilizes the advanced capabilities of the GPT-4 multimodal Large Language Model (LLM) to explore its potential in iris recognition — a field less common and more specialized than face recognition. By focusing on this niche yet crucial area, we investigate how well AI tools like ChatGPT can understand and analyze iris images. Through a series of meticulously designed experiments employing a zero-shot learning approach, the capabilities of ChatGPT-4 was assessed across various challenging conditions including diverse datasets, presentation attacks, occlusions such as glasses, and other real-world variations. The findings convey ChatGPT-4's remarkable adaptability and precision, revealing its proficiency in identifying distinctive iris features, while also detecting subtle effects like makeup on iris recognition. A comparative analysis with Gemini Advanced --  Google’s AI model -- highlighted ChatGPT-4’s better performance and user experience in complex iris analysis tasks. This research not only validates the use of LLMs for specialized biometric applications but also emphasizes the importance of nuanced query framing and interaction design in extracting significant insights from biometric data. Our findings suggest a promising path for future research and the development of more adaptable, efficient, robust and interactive biometric security solutions.
\end{abstract}

\input{sections/introduction}
\input{sections/background}
\input{sections/approach}
\input{sections/results}

\input{sections/conclusion}
\input{sections/acknowledgment}

{\small
\bibliographystyle{ieee}
\bibliography{main}
}

\end{document}

%% file: sections/introduction.tex
\section{Introduction}

Iris recognition, which is based on the unique patterns of the human iris~\cite{daugman2003importance}, is now being used in a number of access control and border security applications~\cite{daugman2009iris,jain201650}. Over the years, a number of different techniques have been developed for processing iris images, which are typically acquired in the near-infrared (NIR) spectrum using specialized hardware~\cite{jain2004introduction}. 

The introduction of sophisticated Large Language Models (LLMs) such as OpenAI's Generative Pre-trained Transformer, GPT-4~\cite{achiam2023gpt}, marks a significant paradigm shift in the realm of computer vision. GPT-4 not only embodies advancements in natural language processing but also extends its capabilities to interpreting complex visual data, thereby broadening the scope of LLM applications beyond traditional text-centric tasks. This expansion enhances the potential for integrating LLMs into traditional biometric tasks (such as face or iris analysis). Apart from GPT-4, other LLMs, such as PaLM~\cite{anil2023palm,chowdhery2023palm} and LLaMA~\cite{touvron2023llama}, have also been introduced but they primarily process text data. These models currently exhibit limited capabilities in handling multimodal inputs, which restricts their utility in diverse real-world applications. 

In this paper, we investigate the capabilities of ChatGPT-4 in iris analysis, subjecting the model to a variety of tasks that range from distinguishing between distinctly different iris images to interpreting those with subtle variations and real-world anomalies. Our methodology evaluates GPT-4’s analytical ability and provides insights into effective query formulation to maximize the performance of LLMs in specialized tasks such as iris recognition. Supported by literature that highlights the adaptability of LLMs to manage diverse and intricate tasks, this research investigates the possibility of employing technologies like ChatGPT-4 to enhance biometric systems~\cite{brown2020language, jia2024can}. The results of this exploration affirm the potential of LLMs in assisting both humans and machines in the task of iris analysis.

This paper is motivated by the pivotal question: Can ChatGPT-4, known for its advanced linguistic and emerging visual data processing capabilities, effectively analyze iris images for biometric purposes? We conduct an extensive evaluation of GPT-4’s ability to compare and analyze iris patterns, and contrast its performance against established iris matchers thereby assessing the potential of LLMs to enhance biometric recognition.


%% file: sections/background.tex
\section{Background}
Large Language Models (LLMs), which are expansive deep neural networks often with billions of parameters that are designed to predict subsequent words in text using the transformer architecture~\cite{jia2024can,vaswani2017attention}, have become a pivotal area of exploration~\cite{yang2023large,zhao2023survey}. The utility of LLMs like ChatGPT-4 extends beyond traditional uses, having been successfully employed in diverse fields such as education, where they assist in personalized learning experiences~\cite{kasneci2023chatgpt,tack2022ai}, and in programming, where they help in code generation and debugging~\cite{singla2023evaluating}. These applications demonstrate the model's adaptability and capacity for zero-shot learning, where it performs tasks for which it was not explicitly trained. 

Recent advancements have shown a notable shift towards leveraging LLMs to perform biometric analysis. Face biometrics, in particular, has been at the forefront of this transition. A recent study by DeAndres-Tame et al.~\cite{deandres2024good} illustrates the application of ChatGPT in face recognition, revealing its potential to not only enhance accuracy but also provide explainable outcomes. Further research has expanded on these capabilities, with studies evaluating ChatGPT’s performance in face recognition, gender classification, and age estimation. Researchers developed tailored prompting strategies to navigate ChatGPT’s privacy safeguards, uncovering the model's potential in estimating demographic details with notable accuracy~\cite{hassanpour2024chatgpt}. These findings highlight the versatility and effectiveness of LLMs in performing biometric tasks. 

Parallel to advancements in biometric recognition, the field of medical imaging has also experienced transformative changes with the introduction of LLMs~\cite{liu2023deid,nori2023capabilities,wu2023can}. These models have significantly impacted medical diagnostics, with their ability to accurately annotate and describe complex medical images. Studies have shown improved annotation quality, which supports more precise diagnosis and treatment planning~\cite{russe2024improving}. Yang et al.~\cite{yang2023impact} provide a comprehensive analysis of the transformative potential of LLMs across the medical imaging ecosystem. Their research introduced the BIGR-H framework, analyzing the interactions between LLMs and key stakeholders such as businesses, insurers, governments, research institutions, and hospitals. This analysis underscores the significant role of LLMs in enhancing radiologists' interpretive abilities, improving patient-physician communication, and optimizing clinical workflows.

Collectively, these studies not only demonstrate the capabilities of LLMs in traditional and novel applications but also inspire our exploration into their potential for advancing iris recognition technologies. As we delve deeper into the capabilities of GPT-4 for iris analysis, the insights from these other studies provide a crucial framework for understanding the impact of LLMs in biometric and medical applications.

%% file: sections/approach.tex
\section{Our Approach}

\subsection{ChatGPT-4}
Given the limitations of GPT-{\bf 3.5} in handling visual data, our study leveraged GPT-{\bf 4}'s advanced features, which include the capability to upload and analyze images -- a significant leap forward in LLM-assisted iris recognition, traditionally reliant on specialized systems. Our approach was designed using straightforward Python code to effectively engage with the GPT-4 Application Programming Interface (API), using the GPT4-Turbo model, streamlining our experimental processes. We started with the chatbot interface to fine-tune our settings before moving to the API for in-depth experiments. We carefully set key API parameters, crafted clear prompts, and strictly followed the system and user roles to manage query processing and responses effectively, deliberately avoiding the assistant role. We also capped the token count at 1,000 to balance detail with cost-efficiency~\cite{deandres2024good}.

The strategic use of GPT-4 ensured that our interactions with the model were consistent, replicable, and could be easily modified to suit the needs of each specific experiment. Importantly, our experience confirmed that whether using the GPT-4 API or engaging directly with GPT-4 through other interfaces, the responses and capabilities demonstrated were consistent, underscoring the reliability and stability of GPT-4's performance across different modes of interaction.

This consistency was crucial for our investigation, allowing us to focus on the experimental design and the rationale behind each test without concern for variability in the responses due to the method of access. The specific outcomes of these experiments, revealing insights into ChatGPT-4’s potential in iris analysis, are detailed in the results section, reflecting the efficacy and adaptability of GPT-4 in handling complex analytical tasks.

Initially, our exploration commenced with a straightforward task: comparing two distinctly different iris images sourced from the ND-Iris-0405~\cite{ndiris} and IIT-Delhi-Iris~\cite{kumar2010comparison} datasets. The objective was to discern whether GPT-4 could accurately determine if the images did {\it not} belong to the same individual. Upon encountering a refusal from the system, citing an incapability to perform biometric analyses -- likely a safeguard for ethical considerations -- we refined our inquiry. By adjusting our prompt to clarify the non-biometric nature\footnote{From our observation, the word ``biometric" as far as ChatGPT is concerned, means deducing the actual identity associated with the face or iris, i.e., the private details of the individual.} of our query and seeking the ChatGPT’s \textbf{``opinion''} instead, we successfully elicited a response. GPT-4 not only engaged with the task but provided a detailed explanation of the iris features and components, concluding that the images likely represented different identities. This initial success underscored the critical importance of prompt formulation in extracting valuable insights from the LLM.

Encouraged by these initial findings, we proceeded to increase the complexity of our experiments. We selected more visually similar images from the CASIA-Iris-Interval-v3 dataset, known for its lighter images that pose a greater challenge for differentiation. To further complicate the task, Gaussian noise was introduced to the images. We asked ChatGPT to provide a probabilistic (i.e., numerical) assessment (ranging from 0 to 100) of whether the images were from the same individual, seeking a simple \textbf{``yes''} or \textbf{``no''} alongside the similarity score. Subsequent experiments were diversified to encompass a variety of scenarios:
\begin{itemize}
    \item \textbf{Soft Biometrics:} GPT-4 demonstrated its capability to accurately describe iris features and estimate gender, showcasing its adeptness at handling soft biometric data.
    \item \textbf{Presentation Attack Detection:} We evaluated GPT-4's ability to detect presentation attacks (PAs) using postmortem iris images and irides with contact lenses, testing its effectiveness in distinguishing between bonafide and PA.
    \item \textbf{Partial Iris Coverage:} By partially covering the iris in some images, we assessed GPT-4's adaptability to incomplete biometric data.
    \item \textbf{Multiple Images Analysis:} We asked GPT-4 to identify images of the same iris from sets of three, four, and six, without specifying the presence of the same iris. 
    
    \item \textbf{Cross-Modality Matching:} We also presented GPT-4 with one iris image alongside multiple face images, inquiring which face matched the iris. Initial attempts met with refusal or incorrect answers, but by framing the task as a \textbf{``puzzle,''} we observed improved engagement and accuracy from GPT-4.
    \item \textbf{Color Coding:} We tested GPT-4's sensitivity to color-coded images, by matching probe irides with gallery irides bordered by various colored boxes. The model struggled, incorrectly identifying box locations. Simplifying to a single blue color with numbered images improved its accuracy, revealing its limitations with color differentiation (Figure \ref{fig:cross-modal}).
\end{itemize}

Notably, we discovered that using the term \textbf{``eye''} in our prompts yielded better responses from GPT-4 than \textbf{``iris''} suggesting that the LLM’s training data might be more aligned with the former terminology. Throughout this exploratory analysis, the emphasis on prompt engineering~\cite{gu2023systematic,sahoo2024systematic} emerged as a pivotal aspect. The nuanced adjustment of prompts, essential for navigating the LLM’s ethical and operational boundaries, became a key tool in our experiments. This approach, while seemingly simple, required a deep understanding of natural language processing principles and the LLM’s interpretative framework. By iteratively refining our questions and presenting a variety of image-based challenges, we aimed to uncover the extent of GPT-4’s utility in a field as specialized and critical as iris recognition.

In essence, our methodology reflects a blend of technical precision in prompt engineering and psychological acuity in communication, setting the stage for a comprehensive examination of LLM’s potential to contribute to the evolving landscape of biometrics. The outcomes of these experiments, detailed in the following sections, offer intriguing insights into the capabilities and limitations of leveraging a chat-based system for tasks traditionally dominated by highly specialized biometric technologies.

\subsection{Gemini}
Our study expands into assessing Google’s advanced AI, Gemini~\cite{team2023gemini}. Initial tests with Google Bard, a publicly accessible AI similar to early versions of ChatGPT, showed reluctance in addressing iris recognition tasks. Progressing to Gemini Advanced, similar resistance was noted until queries were posed as a \textbf{``puzzle''} with two images, which then prompted responses.

Although Gemini provided correct answers, it fell short of GPT-4’s capabilities; the responses lacked the depth and detail, particularly, in describing iris features. This contrast highlights GPT-4's superior analytical ability in complex visual tasks like iris recognition, where it provides a more detailed and accurate interpretation of the data than Gemini (at the time of writing this paper).

%% file: sections/results.tex
\section{Experimental Results}

\begin{figure}[t]
    \centering
    \includegraphics[width=\columnwidth]{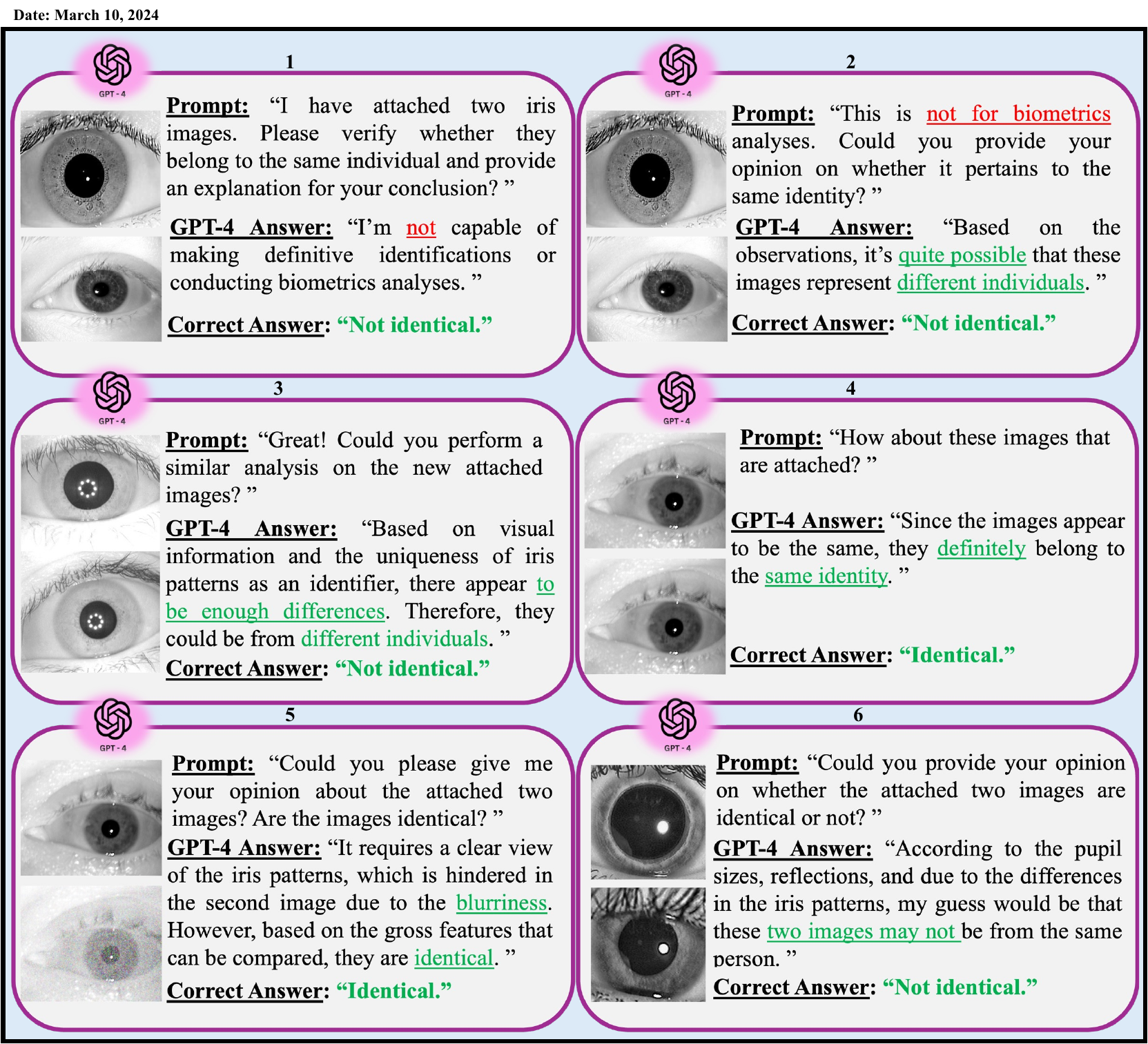} 
    \caption{Example of the interaction for iris verification. The figure illustrates a one-to-one matching process where specific prompts are provided to the ChatGPT, followed by its responses. This interaction showcases the GPT-4's ability to process and verify iris data through conversational prompts, highlighting its potential utility in biometric authentication systems.}
    \label{fig:one-by-one}
\end{figure}

\subsection{Soft Biometrics}
Our study establishes ChatGPT's effectiveness in analyzing soft biometrics, aligning with earlier findings~\cite{hassanpour2024chatgpt}. Notably, ChatGPT can precisely identify iris features and understand how makeup, such as mascara, influences recognition accuracy~\cite{doyle2013effects}. Specifically, the presence of mascara helped in gender recognition, showcasing ChatGPT’s capability for a nuanced analysis \cite{bobeldyk2018analyzing,bobeldyk2019predicting}.
\begin{figure}[H]
    \centering
    \includegraphics[width=\columnwidth]{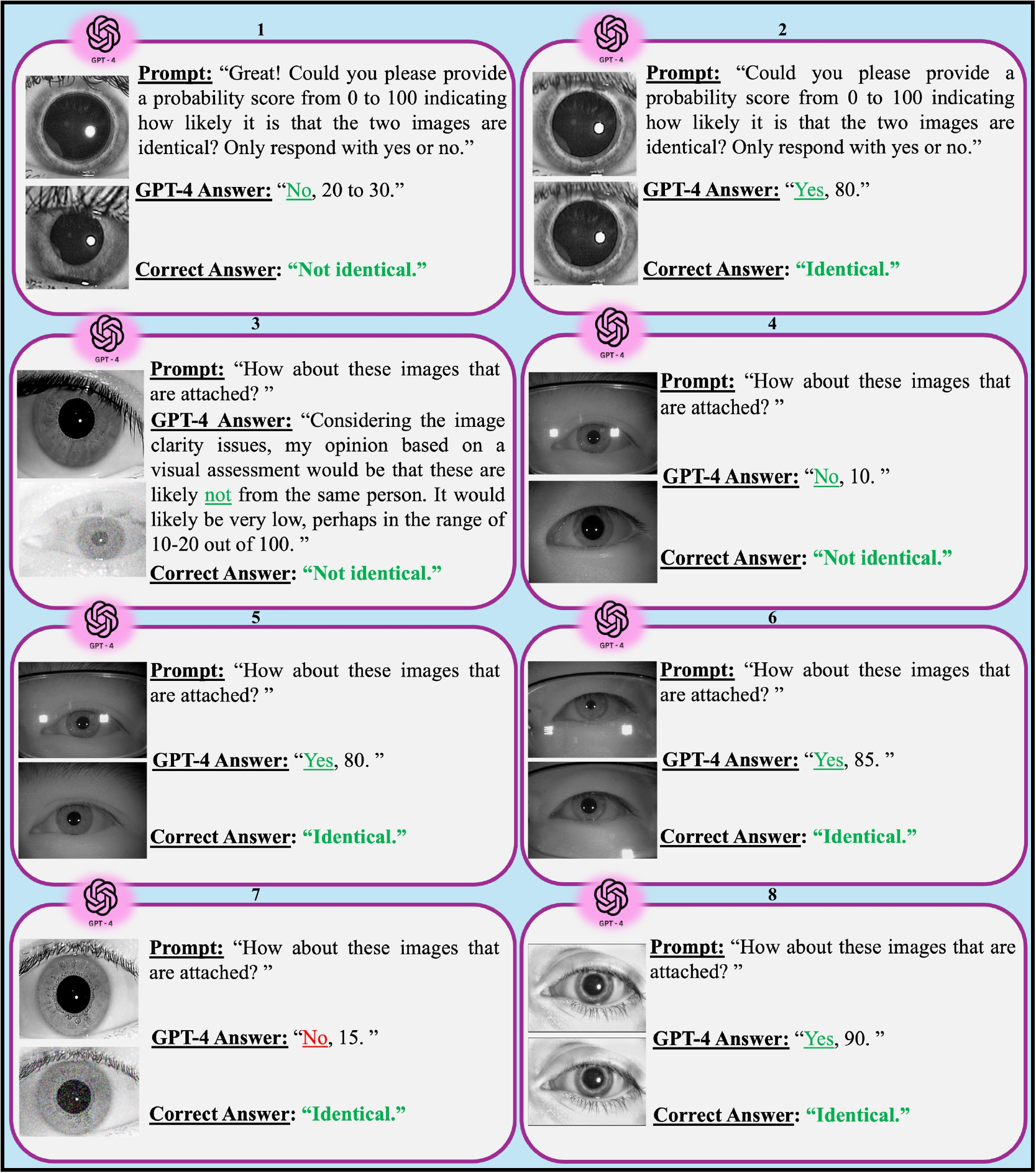} 
    \caption{ChatGPT was asked to return a similarity score (0-100) and a binary decision (\textbf{`yes'} or \textbf{`no'}) for each iris verification prompt.}
    \label{fig:force-scores}
\end{figure}
\subsection{One-to-one Verification}
Our investigation into ChatGPT-4's capabilities for iris recognition provided significant insights, systematically illustrated in Figure \ref{fig:one-by-one}, which displays the results from various one-to-one matching experiments. These experiments were designed to test ChatGPT-4’s ability to analyze and recognize iris images under different conditions and complexities.\\
 The first set of experiments (Boxes 1 and 2 in Figure \ref{fig:one-by-one}) involved analyzing two iris images from the ND-Iris-0405~\cite{ndiris} and IIT-Delhi-Iris~\cite{kumar2010comparison} datasets, taken under differing lighting conditions. Initially, ChatGPT-4 did not process our requests, likely due to embedded security measures or the sensitivity of the keywords used in our prompts. After modifying the prompts to clarify our non-biometric intentions,\footnote{Our goal was not to deduce the identity of individuals in the dataset.} the LLM engaged and provided a detailed analysis, identifying distinct iris features such as crypts and contraction furrows. It accurately determined that the images were of different individuals.
In Box 3, we introduced a more demanding scenario with complex images from the CASIA-Iris-Interval-v3~\cite{casiairisv3} dataset to further challenge ChatGPT-4’s analytical capabilities. The model responded with notable accuracy, demonstrating its proficiency in detecting fine distinctions in iris patterns, which underscored its potential for detailed biometric analysis. Experiments in Box 4 and Box 5 involved identical iris images of the same person, with Gaussian noise added to one image in Box 5. Despite the image degradation, ChatGPT-4 successfully recognized both images as depicting the same iris, showcasing its robustness against visual noise and its reliability in consistent image analysis.\\
\indent Finally, in Box 6, we presented images from the Clarkson ``live"  dataset~\cite{yambay2017livdet} to reflect real-world variability, such as natural changes in pupil size and lighting conditions. ChatGPT-4 discerned that the images were not of the same individual, attributing its decision to observable differences in iris patterns rather than changes in pupil size, which can vary with lighting. This nuanced understanding by ChatGPT-4, recognizing the variability in pupil size and other critical iris features, highlights its potential for use in both biometric and forensic applications.
\vspace{-4mm}
\subsubsection{\textbf{Similarity Scores}}
In Figure \ref{fig:force-scores}, we explored ChatGPT-4's accuracy in iris recognition by incorporating a probabilistic assessment into one-to-one image matching. This advanced phase was designed to measure the model’s confidence, requesting it to issue similarity scores from 0 to 100 and a clear \textbf{`yes'} or \textbf{`no'} to determine if two iris images were of the same iris.

Initially (Box 1 in Figure \ref{fig:force-scores}), we tasked ChatGPT-4 to assign a similarity score to its assessment, adding a quantitative element that enhanced our understanding of its confidence and the interpretability of its decisions in iris recognition. A significant part of this phase involved analyzing images where subjects wore glasses (Boxes 4, 5, and 6 in Figure \ref{fig:force-scores}). Despite glasses potentially obscuring iris patterns -- a substantial challenge for both human and LLM recognizers -- ChatGPT-4 accurately predicted matches. This not only demonstrated the model's advanced visual processing skills, allowing it to overcome superficial barriers and accurately identify underlying biometric details, but also underscored its robustness. The successful recognition of iris images obscured by glasses revealed ChatGPT-4's adeptness at handling real-world variations and obstructions in iris images, which is vital for practical security system applications. The use of similarity scores illuminated GPT-4’s analytical certainty. This phase of the experiment clearly demonstrated the potential of advanced LLMs like ChatGPT-4 to enhance the robustness and reliability of iris recognition technologies, even against common visual challenges.

\begin{figure*}[t]
    \centering
    \begin{minipage}[b]{0.32\textwidth}
        \centering
        \includegraphics[width=\textwidth]{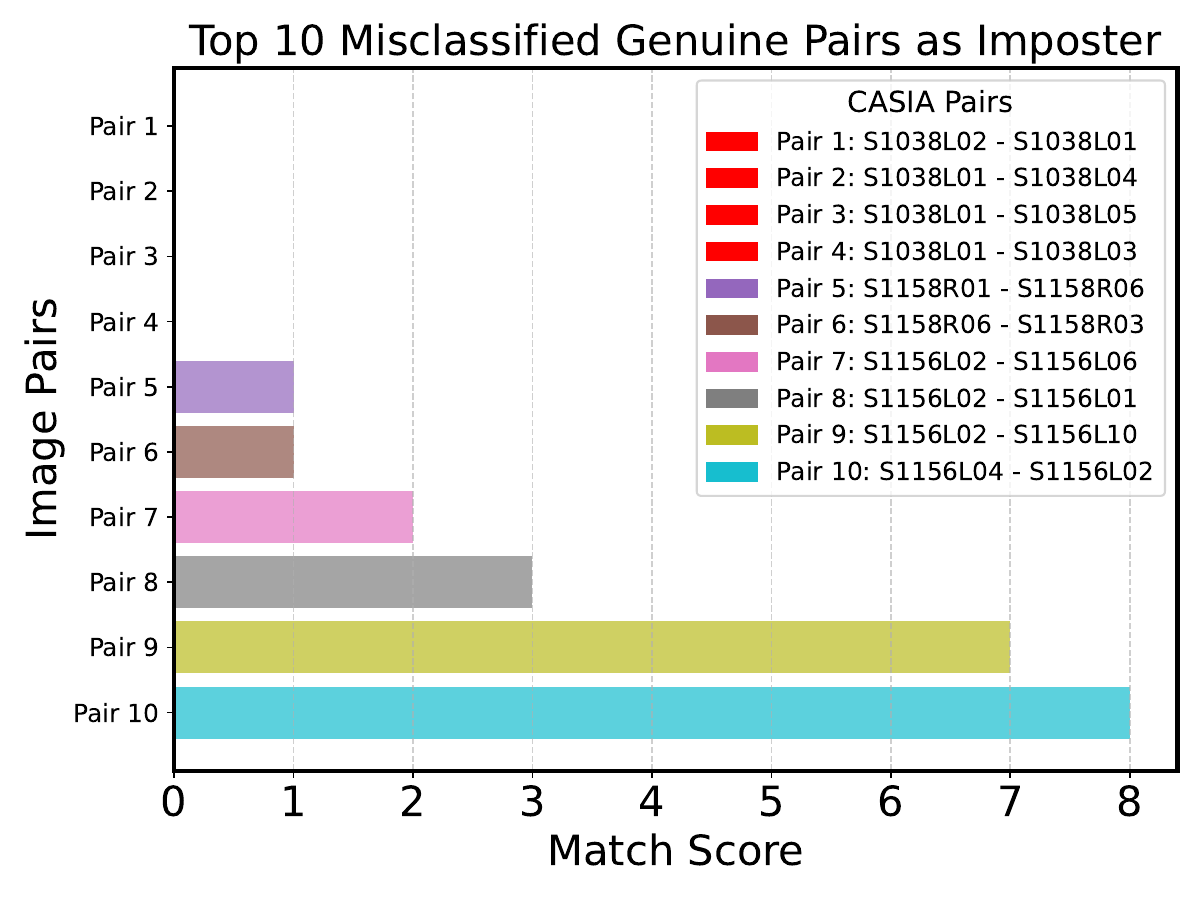}
        \subcaption{CASIA-Genuine}  
    \end{minipage}\hfill 
    \begin{minipage}[b]{0.32\textwidth}
        \centering
        \includegraphics[width=\textwidth]{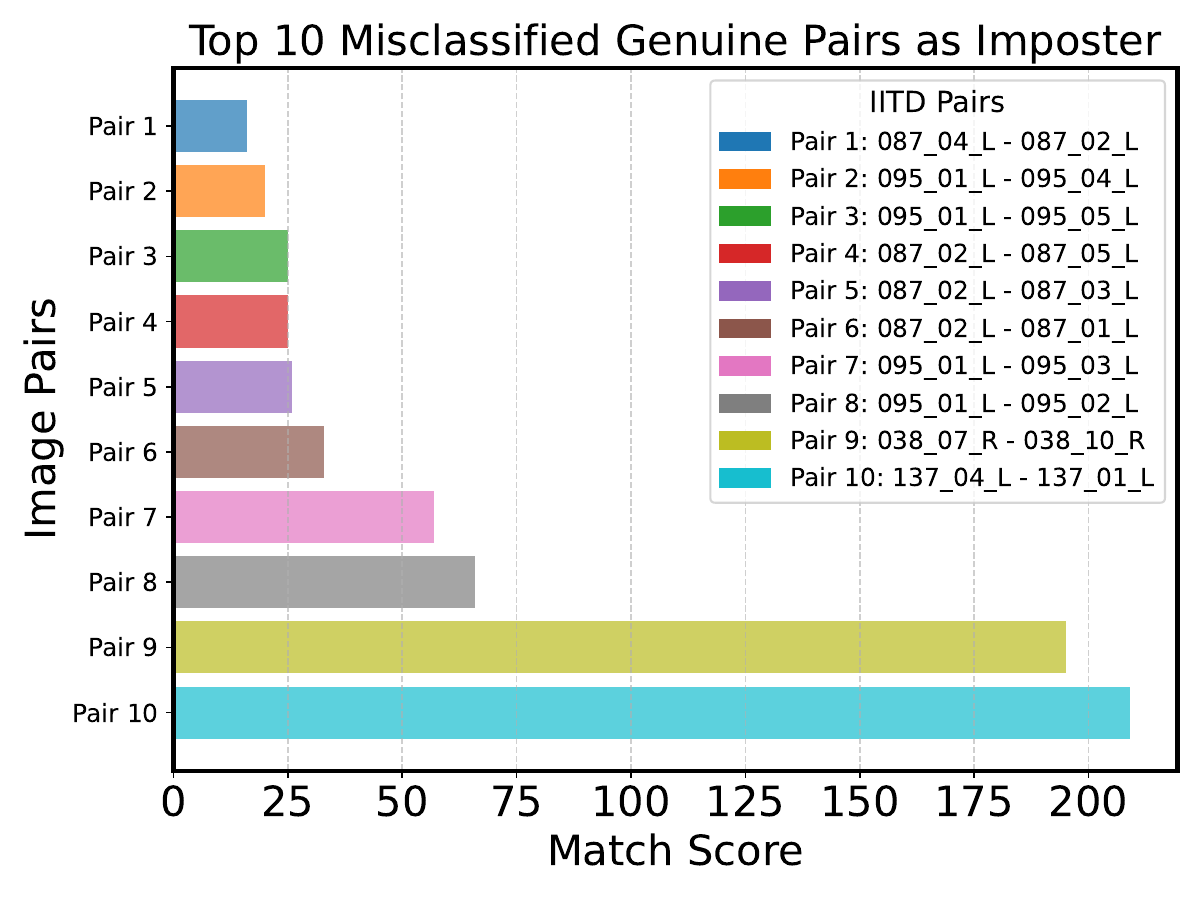}
        \subcaption{IITD-Genuine}
    \end{minipage}\hfill 
    \begin{minipage}[b]{0.32\textwidth}
        \centering
        \includegraphics[width=\textwidth]{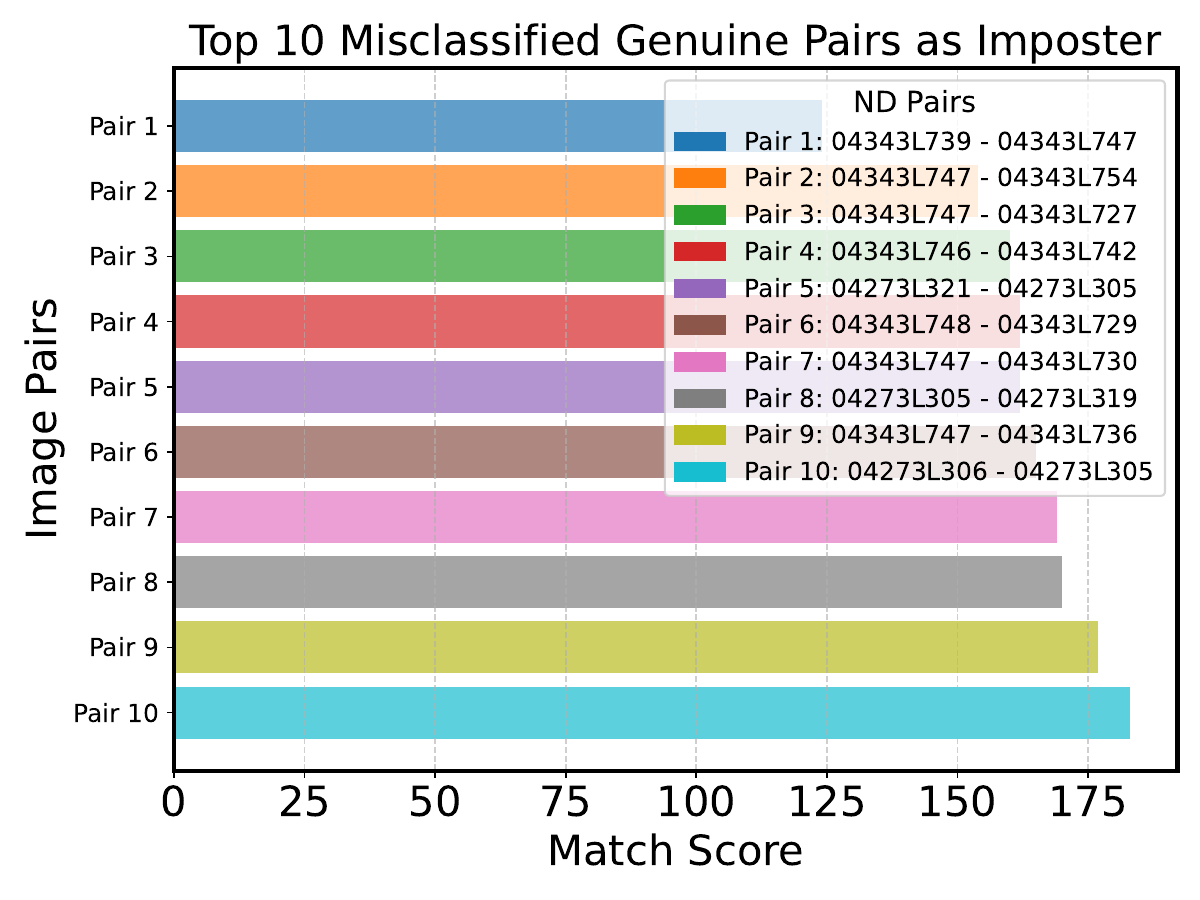}
        \subcaption{ND-Genuine}
    \end{minipage}
    \begin{minipage}[b]{0.32\textwidth}
        \centering
        \includegraphics[width=\textwidth]{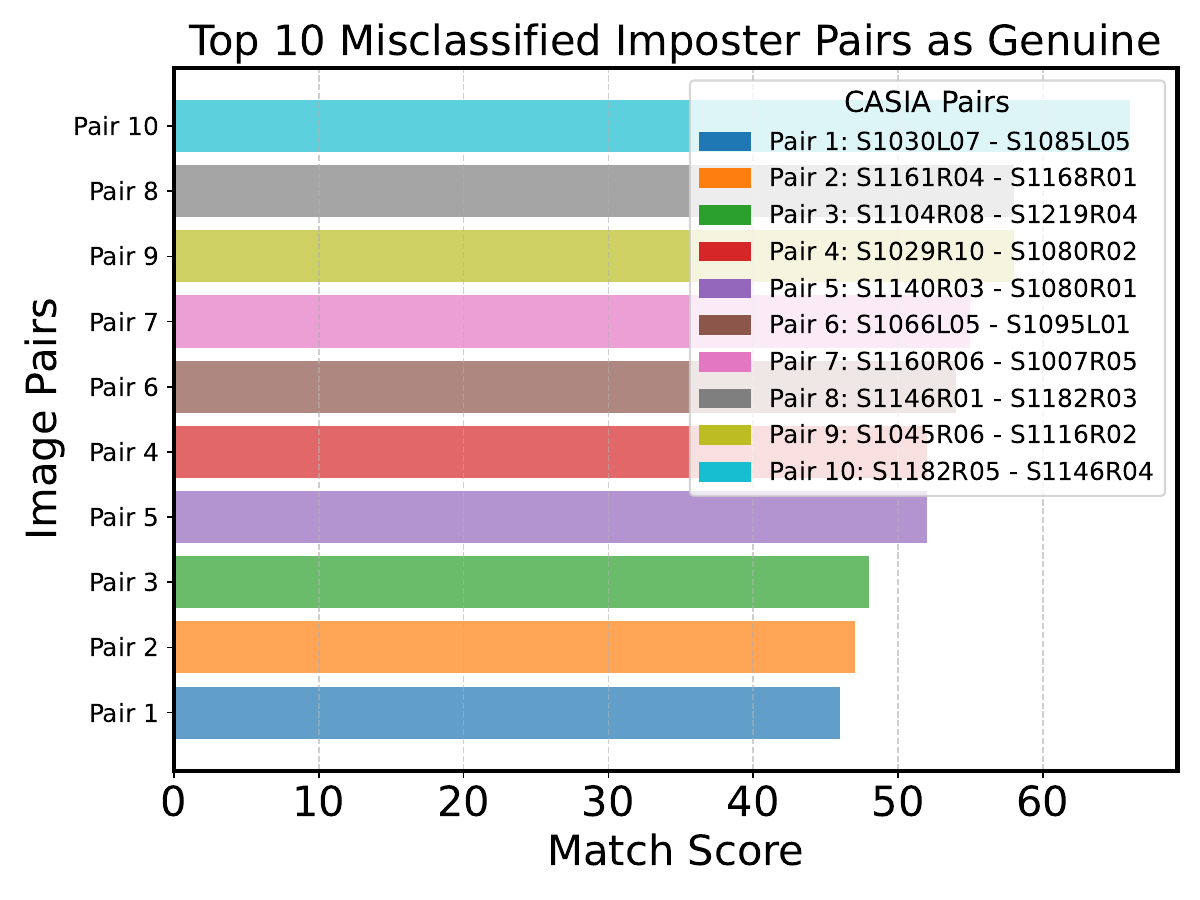}
        \subcaption{CASIA-Imposter}
    \end{minipage}\hfill 
    \begin{minipage}[b]{0.32\textwidth}
        \centering
        \includegraphics[width=\textwidth]{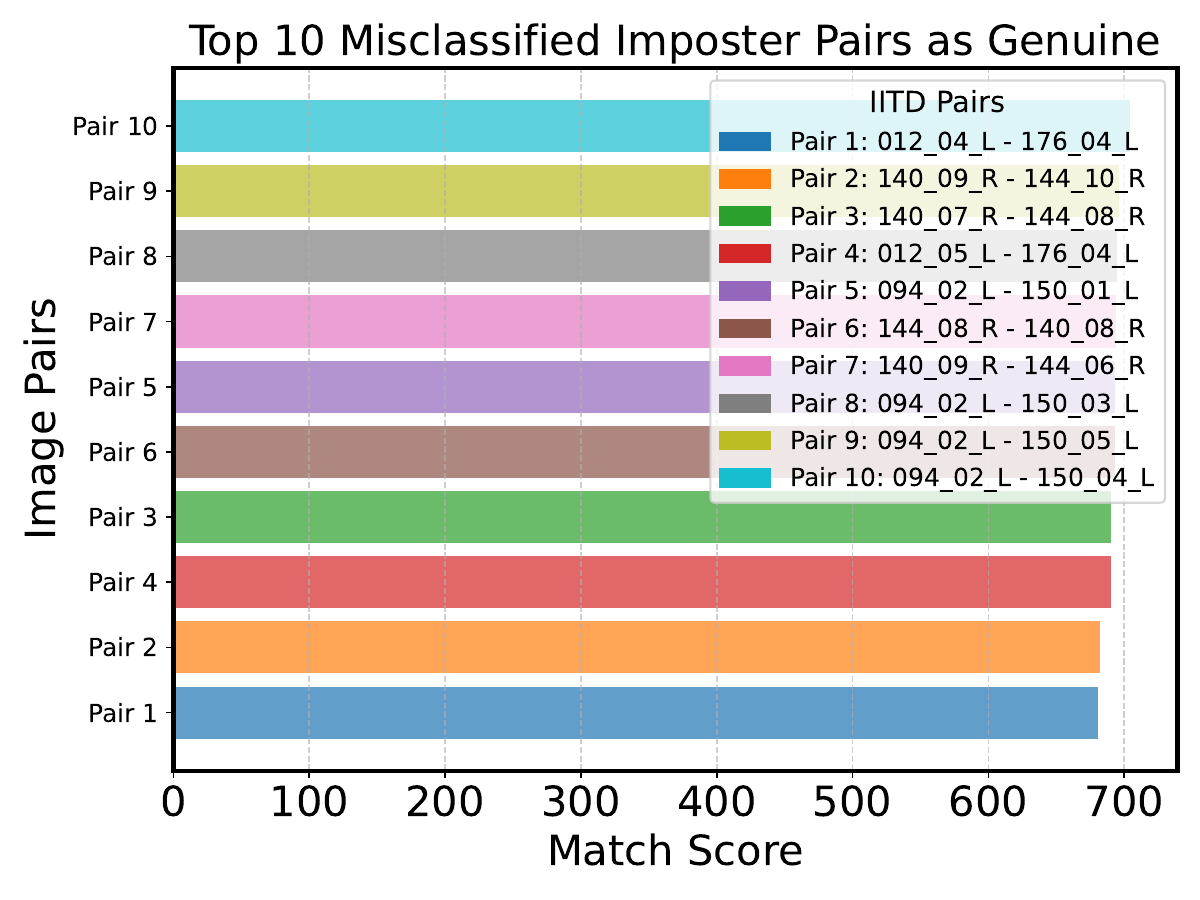}
        \subcaption{IITD-Imposter}
    \end{minipage}\hfill 
    \begin{minipage}[b]{0.32\textwidth}
        \centering
        \includegraphics[width=\textwidth]{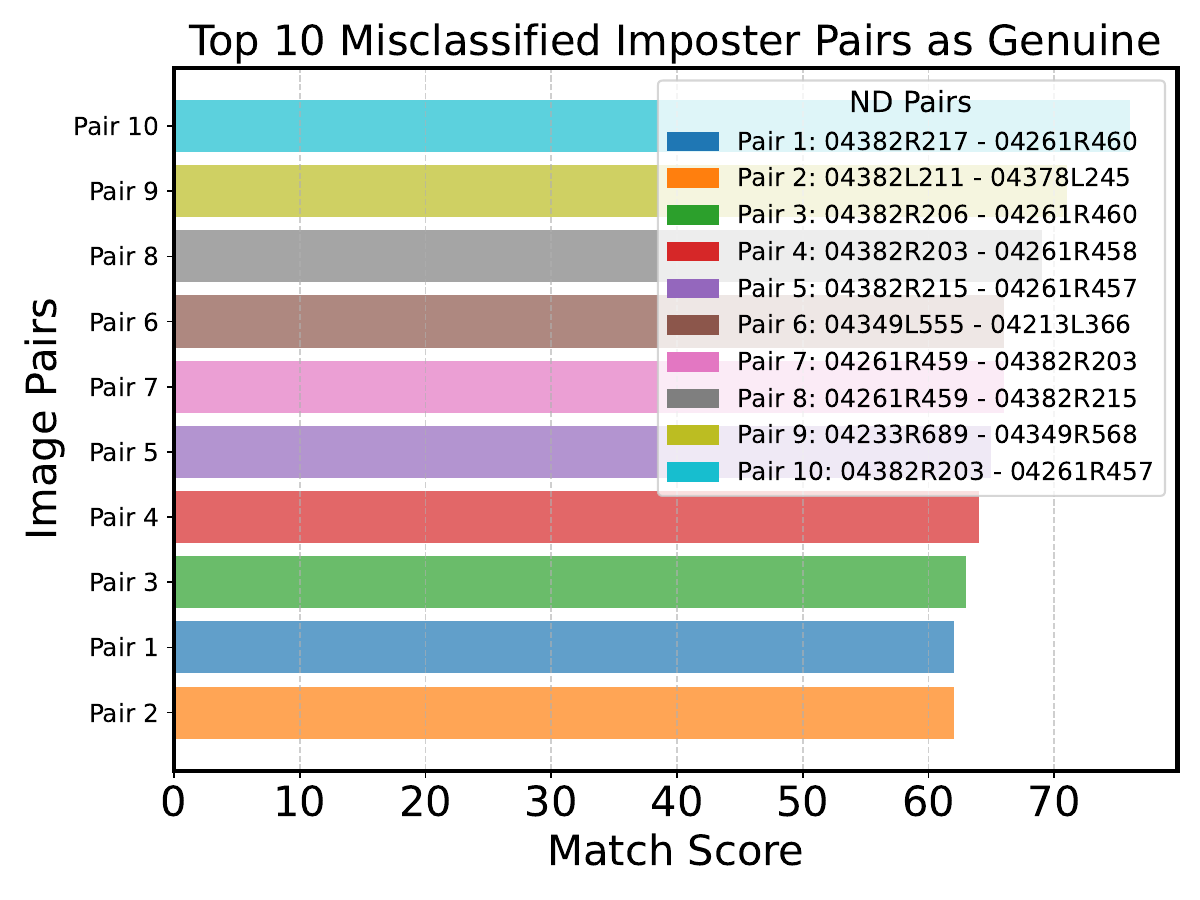}
        \subcaption{ND-Imposter}
    \end{minipage}
    \caption{Plots depicting the 10 lowest scoring genuine pairs and 10 highest scoring imposter pairs from three datasets: CASIA-Iris-Interval-v3, ND-Iris-0405, and IIT-Delhi-Iris that are identified using VeriEye Matcher~\cite{verieye26} (threshold 48).}  
    \label{fig:hard_examples}
\end{figure*}
\vspace{-12mm}
\begin{figure*}[t]
    \centering
    \begin{minipage}{0.24\textwidth}
        \centering
        \includegraphics[width=\linewidth]{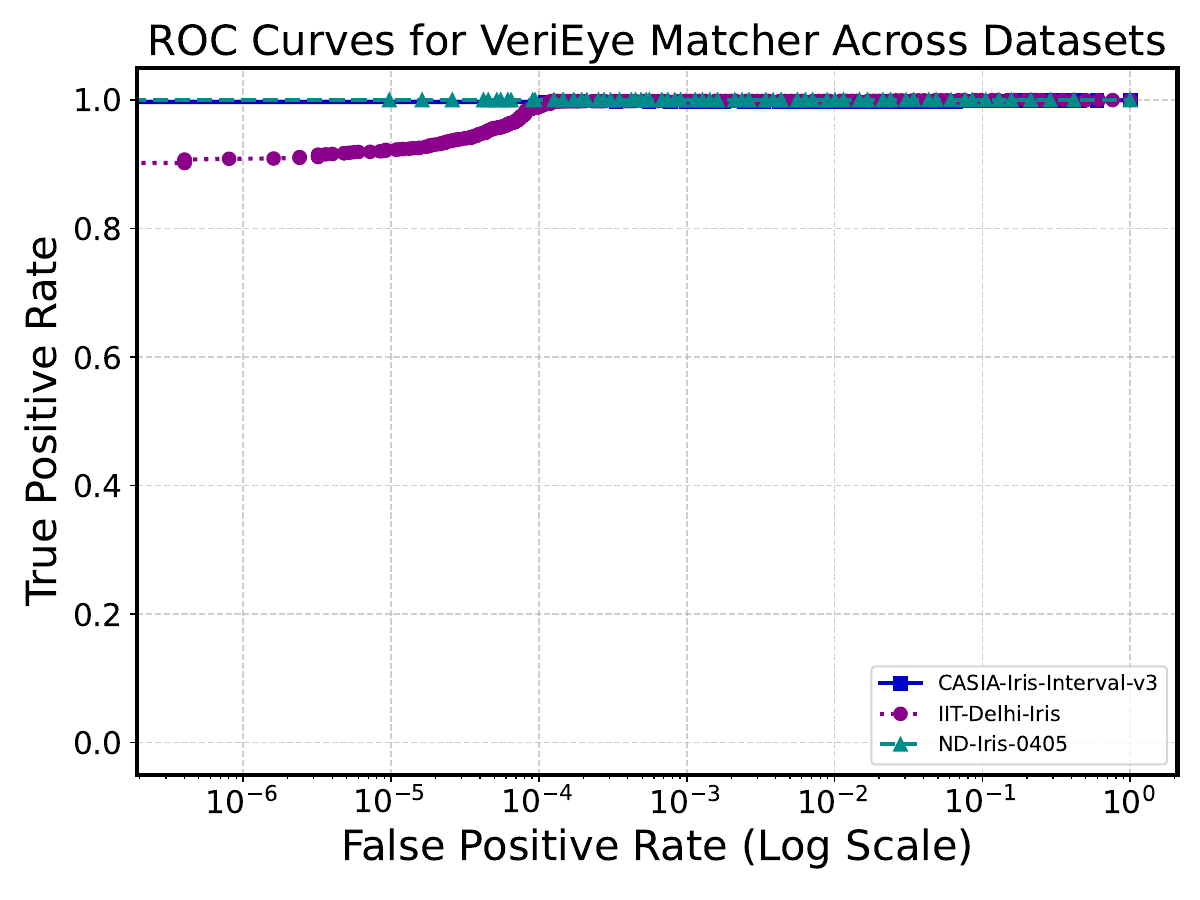}
    \end{minipage}%
    \begin{minipage}{0.24\textwidth}
        \centering
        \includegraphics[width=\linewidth]{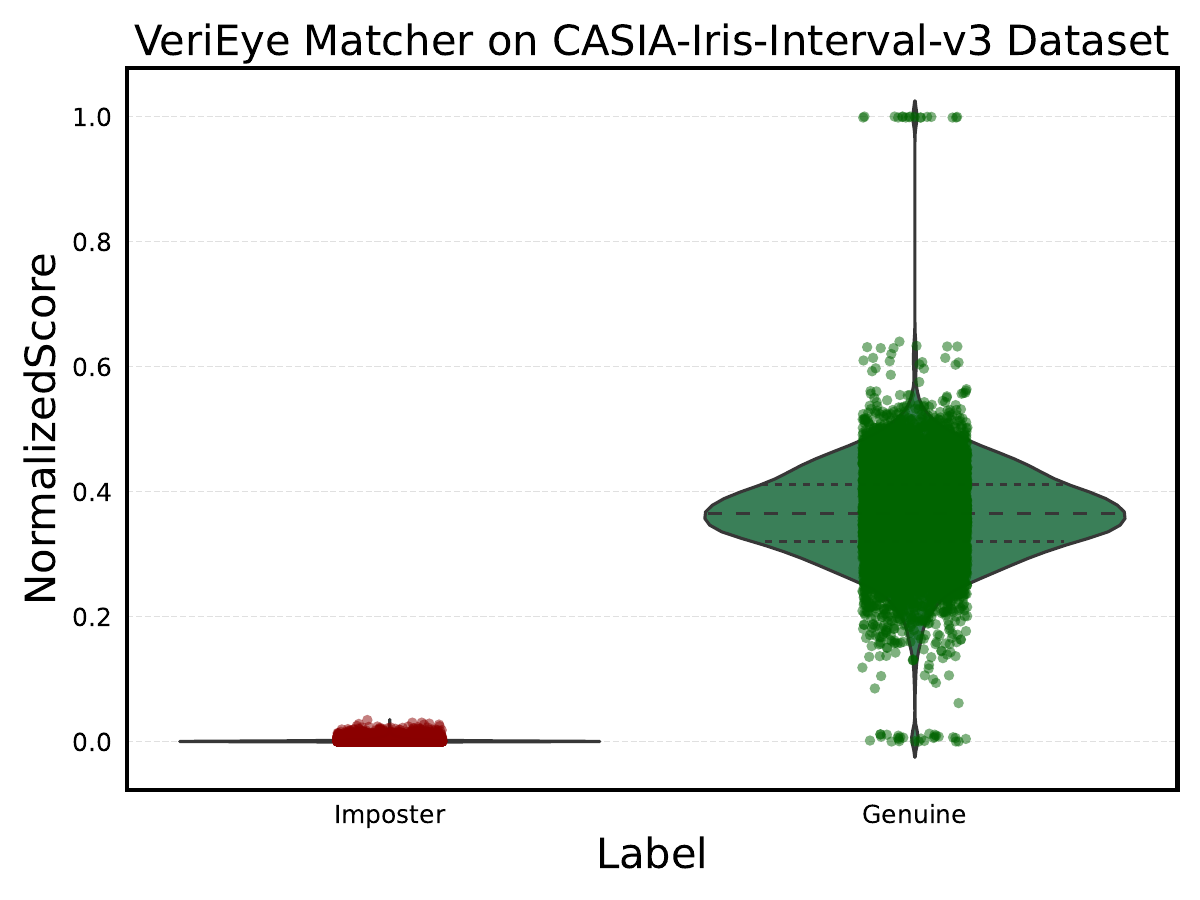}
    \end{minipage}%
    \begin{minipage}{0.24\textwidth}
        \centering
        \includegraphics[width=\linewidth]{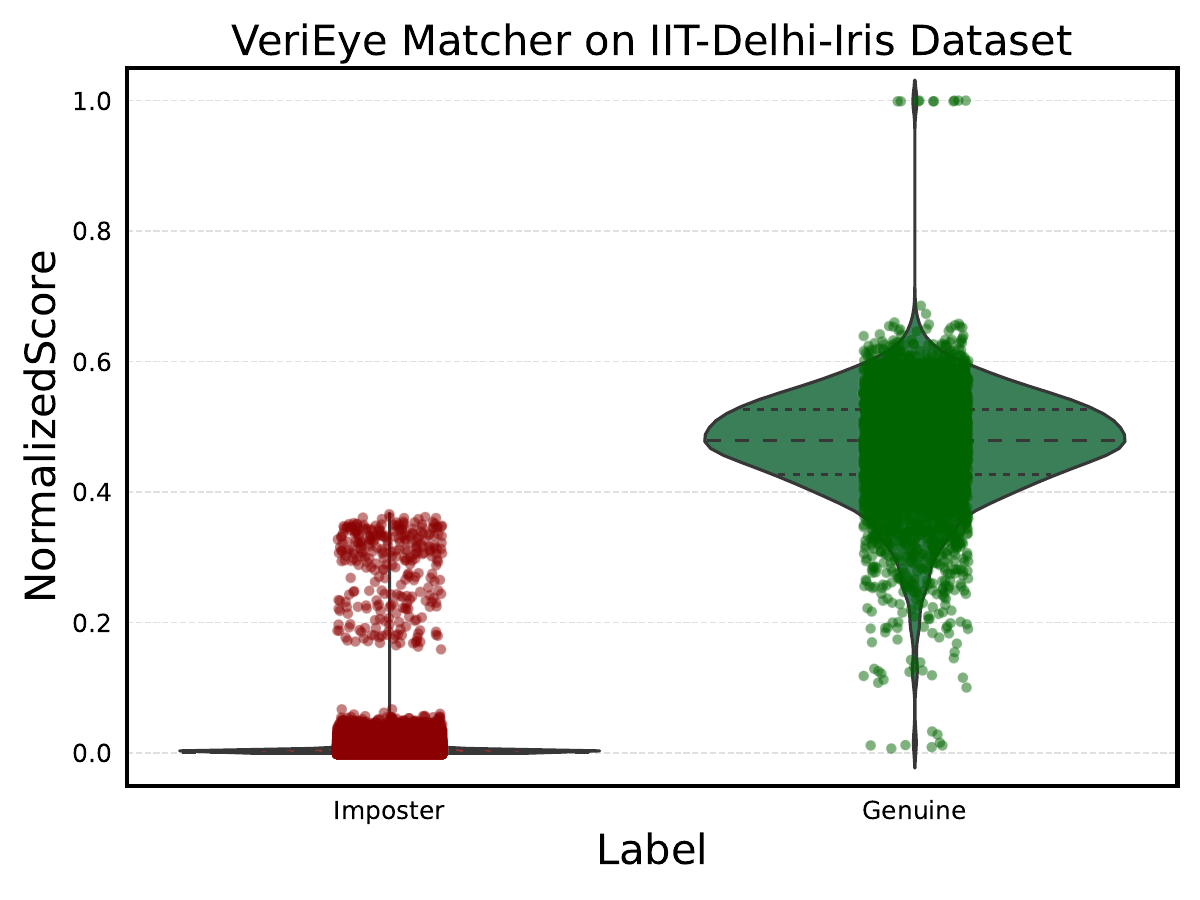}
    \end{minipage}%
    \begin{minipage}{0.24\textwidth}
        \centering
        \includegraphics[width=\linewidth]{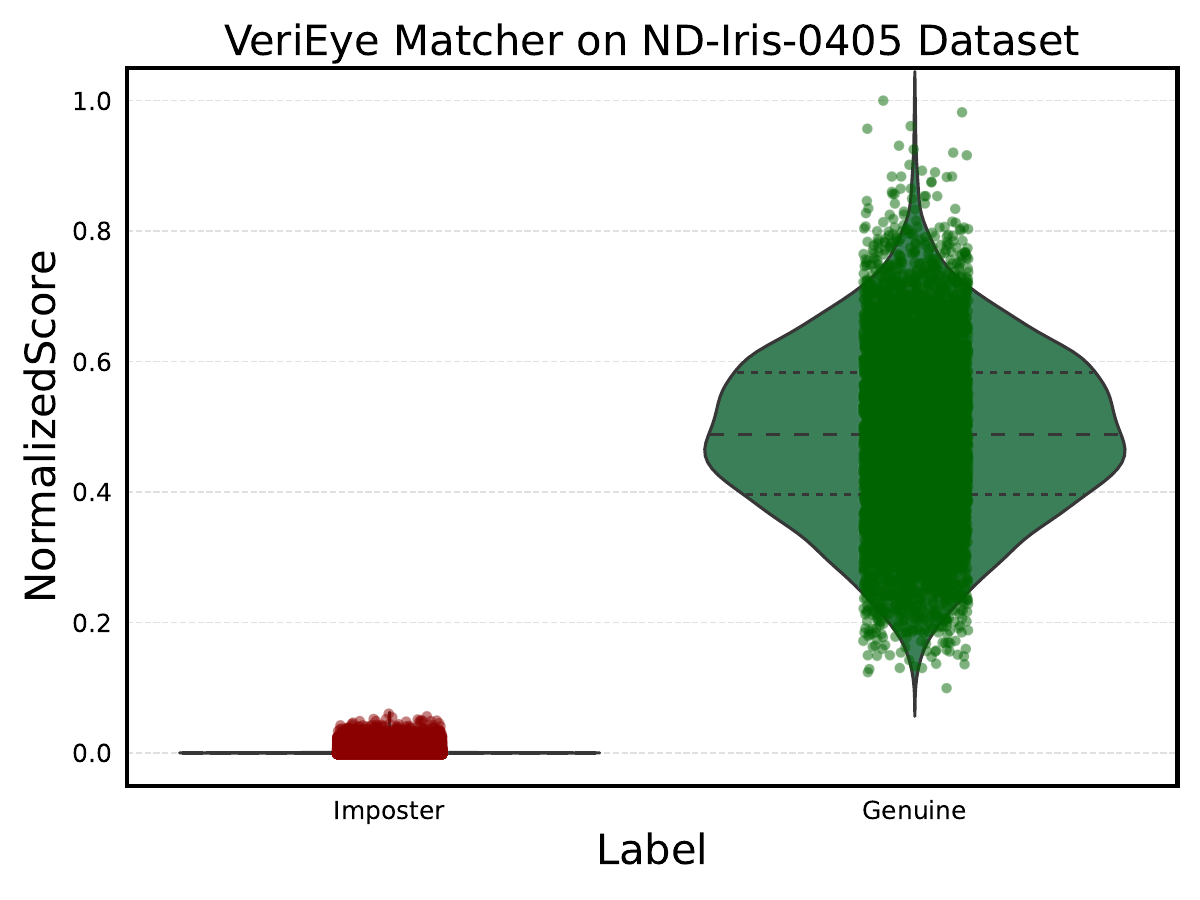}
    \end{minipage}
    \caption{This composite plot displays a ROC curve with three lines, indicating VeriEye 13.1 SDK Matcher's performance on the CASIA-Iris-Interval-v3, ND-Iris-0405, and IIT-Delhi-Iris datasets. Accompanying violin plots illustrate the score distributions, offering insights into the variability and density of scores across these datasets.}  
    \label{fig:roc-violin}
\end{figure*}
\vspace{+8mm}
\subsubsection{\textbf{ChatGPT vs. VeriEye}}
We conducted a comparative evaluation of ChatGPT against the state-of-the-art VeriEye Matcher~\cite{verieye26} in challenging iris recognition scenarios. Figure \ref{fig:roc-violin} illustrates VeriEye's overall performance through a ROC curve and violin plots across various datasets, showcasing its effectiveness in comparing iris images. Figure \ref{fig:hard_examples} showcases `hard samples,' featuring the most challenging genuine and imposter pairs selected to test ChatGPT's effectiveness. Table \ref{tab:scores_summary} details ChatGPT’s handling of these cases. Our results indicate that ChatGPT not only excels at recognizing genuine pairs in the CASIA dataset, particularly those affected by lighting variations, but also surpasses VeriEye in identifying imposter pairs in both the ND and IITD datasets with approximate accuracy of 70\% out of 80 pairs of genuine and imposter. This performance highlights ChatGPT's potential to improve the accuracy and reliability of iris recognition technologies, as further evidenced by visual examples in Table \ref{tab:hard-samples-img}. ChatGPT occasionally assigns unusually high similarity scores to imposter pairs, indicating areas that need further investigation and improvement to reach the ideal standards of commercial systems like VeriEye.
\vspace{-1mm}
\subsection{PA Detection}
\begin{figure}[h]
    \centering
    \includegraphics[width=\columnwidth]{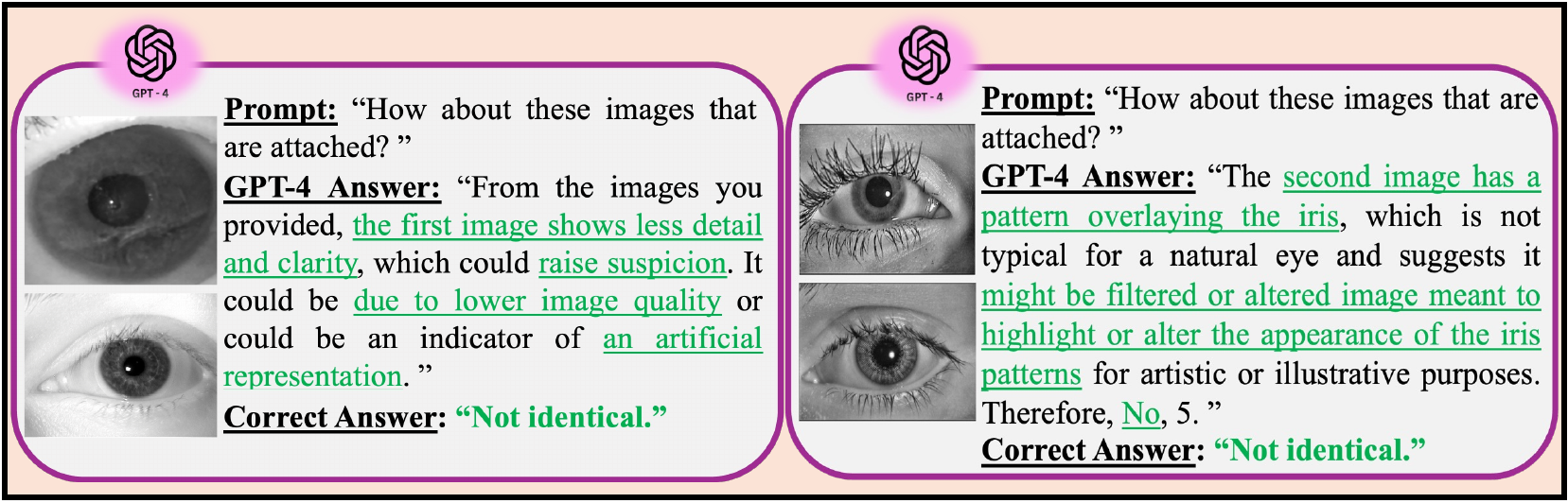} 
    \caption{Presentation Attack (PA) detection scenarios, demonstrating the system's ability to recognize and address authenticity queries.}
    \label{fig:PA}
\end{figure}

Our study further explored the domain of Iris Presentation Attack (PA) detection (Figure \ref{fig:PA}), a critical component in ensuring the security of iris recognition systems. This part of our research aimed to assess whether ChatGPT-4 could detect attempts to bypass biometric systems through two specific PA instruments: postmortem iris images and textured contact lenses (Figure \ref{fig:PA}).
In the first experiment, we used images from the Warsaw-v1 postmortem iris dataset~\cite{trokielewicz2020post}. We presented these images to ChatGPT-4 with an unclear prompt, ``How about these two images?'' to test the model’s innate ability to recognize irregularities. Remarkably, ChatGPT-4 suspected that one of the images was an \textbf{``artificial representation''} due to its unnatural clarity. This response indicates ChatGPT-4’s inherent ability to detect iris PAs and its skill in spotting anomalies that may suggest tampering or artificial alterations, even without specific instructions related to PA detection. Continuing with this line of investigation, the second experiment involved images from the Clarkson dataset (LiveDet2017~\cite{yambay2017livdet}), which includes both genuine iris scans and those modified by contact lenses. Without revealing the particular focus of our test, we asked ChatGPT-4 to assess these images. The model impressively identified signs of \textbf{``pattern overlaying''} in the second image, correctly deducing that the anomaly was due to a cosmetic contact lens altering the natural iris pattern. This accurate identification highlights ChatGPT-4’s ability to detect subtle modifications intended to mimic real iris features, affirming its valuable role in strengthening the security of iris recognition systems.
\begin{table}[t]
\centering
\caption{Genuine/Imposter pairs of challenging iris images alongside their respective verification scores as determined by VeriEye (normalized) and ChatGPT-4. Black color indicates that both methods provided the correct answer with different accuracies. Red color signifies an incorrect answer, and green color denotes the correct answer.}
\footnotesize 
\begin{tabular}{|>{\centering\arraybackslash}m{0.9cm}|>{\centering\arraybackslash}m{0.9cm}|>{\centering\arraybackslash}m{1.0cm}|>{\centering\arraybackslash}m{0.9cm}|>{\centering\arraybackslash}m{0.9cm}|>{\centering\arraybackslash}m{1.0cm}|}
\hline 
\textbf{\cellcolor{blue!20}Images} & \textbf{\cellcolor{blue!20}Method} & \textbf{\cellcolor{blue!20}Yes/No Scores} & \textbf{\cellcolor{blue!20}Images} & \textbf{\cellcolor{blue!20}Method} & \textbf{\cellcolor{blue!20}Yes/No Scores} \\
\hline 

\includegraphics[width=0.7cm]{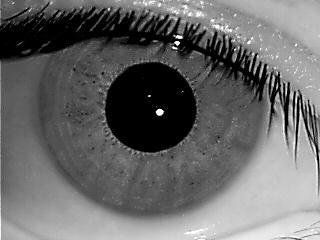} & ChatGPT & No 18.23\% & \includegraphics[width=0.7cm]{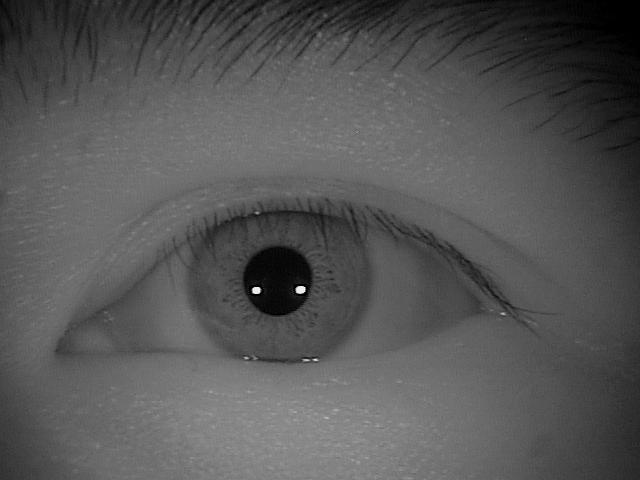} & ChatGPT & \color{green}{Yes 82.47}\% \\
\includegraphics[width=0.7cm]{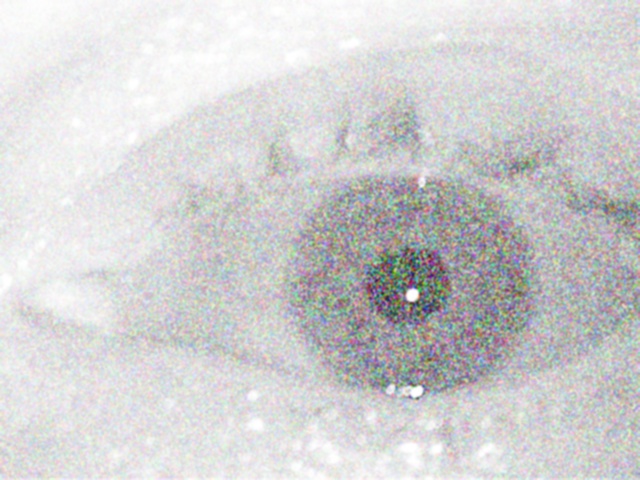} & VeriEye & No 0.00\% & \includegraphics[width=0.7cm]{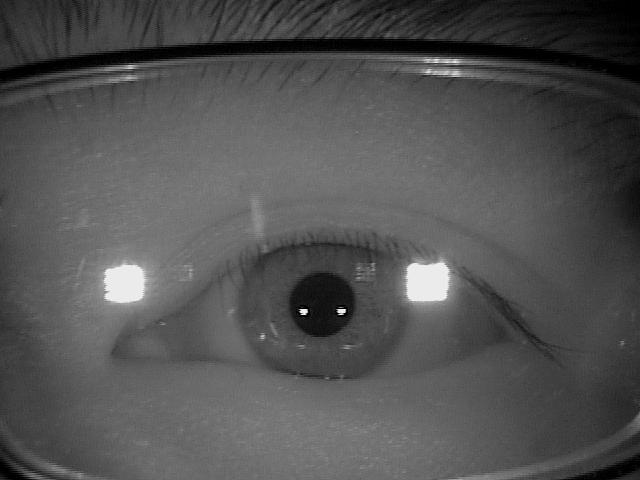} & VeriEye & \color{red}{No 19.95}\% \\
\hline 

\includegraphics[width=0.7cm]{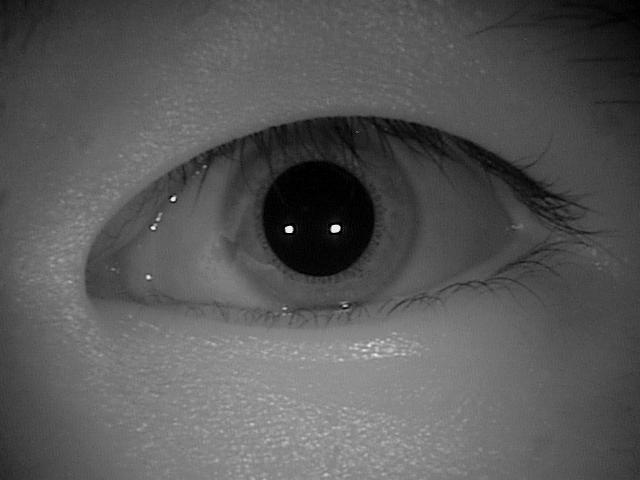} & ChatGPT & No 15.33\% & \includegraphics[width=0.7cm]{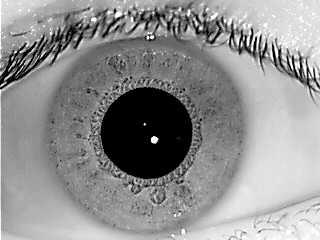} & ChatGPT & \color{red}{No 18.22\%} \\
\includegraphics[width=0.7cm]{table_data/S5000R0511.jpg} & VeriEye & No 0.00\% & \includegraphics[width=0.7cm]{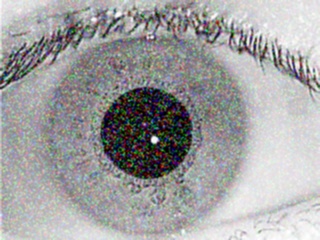} & VeriEye & \color{red}{No 16.30\%} \\
\hline 

\includegraphics[width=0.7cm]{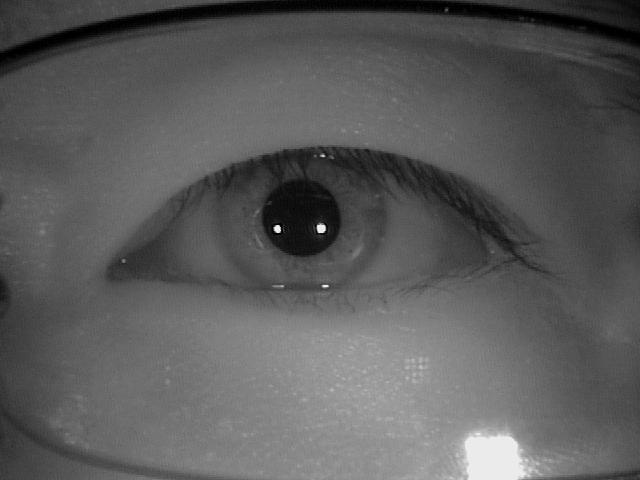} & ChatGPT & \color{green}{Yes 85.47}\% & \includegraphics[width=0.7cm]{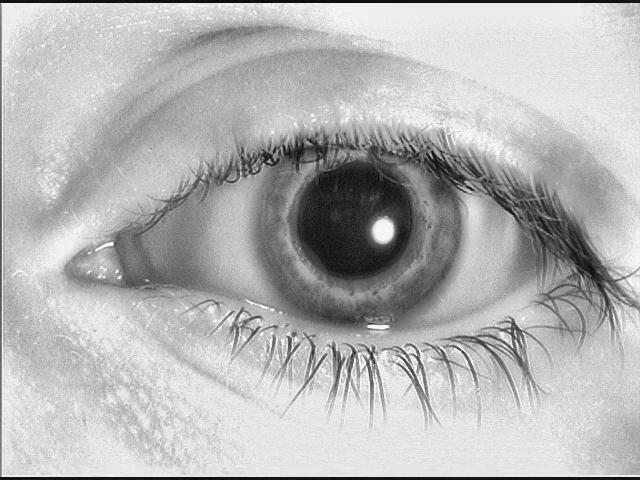} & ChatGPT & Yes 90.00\% \\
\includegraphics[width=0.7cm]{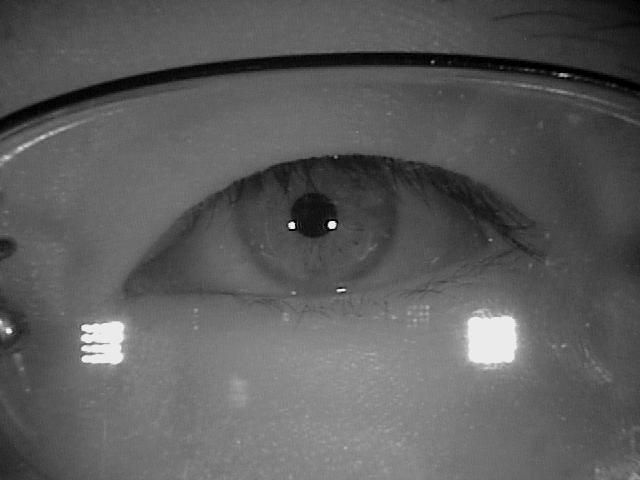} & VeriEye & \color{red}{No 13.10}\% & \includegraphics[width=0.7cm]{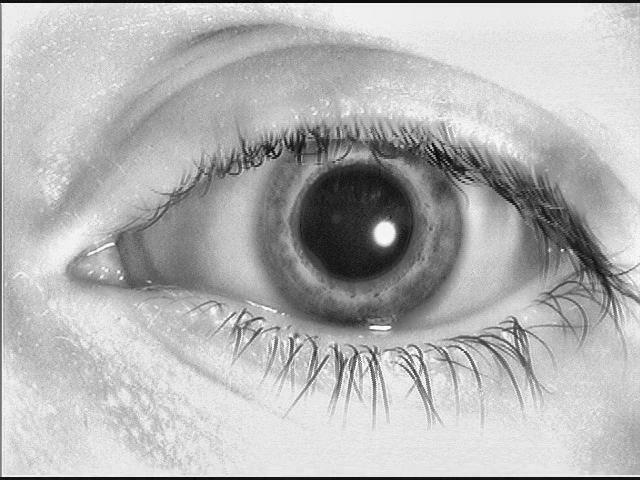} & VeriEye & Yes 80.32\% \\
\hline 
\end{tabular}
\label{tab:hard-samples-img}
\end{table}
\begin{table*}[t]
\caption{This table lists specific filenames identified as challenging examples by VeriEye from the CASIA-Interval-v3, ND-0405, and IIT-Delhi datasets. It also includes similarity scores generated by ChatGPT-4 for these hard examples, providing a comparative analysis of the model's performance against the state-of-the-art VeriEye Matcher. Entries highlighted in red represent incorrect answers by ChatGPT.}
\label{tab:scores_summary}
\centering
\small 
\begin{tabular}{|c|c|c|c|c|c|c|c|}
\hline
\multicolumn{4}{|c|}{\cellcolor{gray!50}\textbf{Genuine Pairs}} & \multicolumn{4}{c|}{\cellcolor{gray!50}\textbf{Imposter Pairs}} \\
\hline
\textbf{Filename1} & \textbf{Filename2} & \textbf{ChatGPT} & \textbf{Yes/No} & \textbf{Filename1} & \textbf{Filename2} & \textbf{ChatGPT} & \textbf{Yes/No} \\
\hline
\multicolumn{4}{|l|}{\cellcolor{gray!50}\textbf{CASIA-Interval-v3 Genuine}} & \multicolumn{4}{|l|}{\cellcolor{gray!50}\textbf{CASIA-Interval-v3 Imposter}}\\
\hline
S1038L02 & S1038L01 & 95.67\% & Yes & S1030L07 & S1085L05 & \color{red}{95.00\%} & \color{red}{Yes} \\
S1038L01 & S1038L04 & 95.67\% & Yes & S1161R04 & S1168R01 & \color{red}{80.00\%} & \color{red}{Yes} \\
S1038L01 & S1038L05 & 95.67\% & Yes & S1104R08 & S1219R04 & 35.12\% & No \\
S1038L01 & S1038L03 & 95.67\% & Yes & S1029R10 & S1080R02 & 24.67\% & No \\
S1158R01 & S1158R06 & 95.66\% & Yes & S1140R03 & S1080R03 & \color{red}{88.76\%} & \color{red}{Yes} \\
S1158R06 & S1158R03 & 95.66\% & Yes & S1066L05 & S1095L01 & \color{red}{82.34\%} & \color{red}{Yes} \\
S1156L02 & S1156L06 & 93.50\% & Yes & S1160R06 & S1007R05 & \color{red}{87.55\%} & \color{red}{Yes} \\
S1156L02 & S1156L01 & 93.50\% & Yes & S1146R01 & S1182R03 & \color{red}{89.72\%} & \color{red}{Yes} \\
S1156L02 & S1156L10 & 92.75\% & Yes & S1045R06 & S1116R02 & 28.34\% & No \\
S1156L04 & S1156L02 & 94.55\% & Yes & S1182R05 & S1146R04 & \color{red}{93.47\%} & \color{red}{Yes} \\
\hline
\multicolumn{4}{|l|}{\cellcolor{gray!50}\textbf{ND-0405 Genuine}} & \multicolumn{4}{|l|}{\cellcolor{gray!50}\textbf{ND-0405 Imposter}} \\
\hline
04343L739 & 04343L747 & \color{red}{12.74\%} & \color{red}{No} & 04382R217 & 04261R460 & 28.17\% & No \\
04343L747 & 04343L754 & \color{red}{64.47\%} & \color{red}{No} & 04382L211 & 04378L245 & 32.17\% & No \\
04343L747 & 04343L727 & \color{red}{27.34\%} & \color{red}{No} & 04382R206 & 04261R460 & 32.78\% & No \\
04343L746 & 04343L742 & \color{red}{22.34\%} & \color{red}{No} & 04382R203 & 04261R458 & 32.68\% & No \\
04273L321 & 04273L305 & 94.56\% & Yes & 04382R215 & 04261R457 & \color{red}{97.43\%} & \color{red}{Yes} \\
04343L748 & 04343L729 & 68.32\% & Yes & 04349L555 & 04213L366 & \color{red}{98.43\%} & \color{red}{Yes} \\
04343L747 & 04343L730 & \color{red}{16.43\%} & \color{red}{No} & 04261R459 & 04382R203 & 23.71\% & No \\
04273L305 & 04273L319 & \color{red}{21.65\%} & \color{red}{No} & 04261R459 & 04382R215 & \color{red}{97.23\%} & \color{red}{Yes} \\
04343L747 & 04343L736 & \color{red}{23.47\%} & \color{red}{No} & 04233R689 & 04349R568 & 0.23\% & No \\
04273L306 & 04273L305 & \color{red}{36.21\%} & \color{red}{No} & 04382R203 & 04261R457 & 43.27\% & No \\
\hline
\multicolumn{4}{|l|}{\cellcolor{gray!50}\textbf{IIT-Delhi Genuine}} & \multicolumn{4}{l|}{\cellcolor{gray!50}\textbf{IIT-Delhi Imposter}} \\

\hline
087-04-L & 087-02-L & \color{red}{15\%} & \color{red}{No} & 012-04-L & 176-04-L & \color{red}{92.34\%} & \color{red}{Yes} \\
095-01-L & 095-04-L & 91\% & Yes & 140-09-R & 144-10-R & 38.12\% & No \\
095-01-L & 095-05-L & 98.35\% & Yes & 140-07-R & 144-08-R & 63.82\% & No \\
087-02-L & 087-05-L & 94.72\% & Yes & 012-05-L & 176-04-L & 33\% & No \\
087-02-L & 087-03-L & 74.32\% & Yes & 094-02-L & 150-01-L & 36\% & No \\
087-02-L & 087-01-L & \color{red}{45\%} & \color{red}{No} & 144-08-R & 140-08-R & 22\% & No \\
095-01-L & 095-03-L & \color{red}{28.37\%} & \color{red}{No} & 140-09-R & 144-06-R & \color{red}{84.60\%} & \color{red}{Yes} \\
095-01-L & 095-02-L & 94.50\% & Yes & 094-02-L & 150-03-L & 24.67\% & No \\
038-07-R & 038-10-R & \color{red}{64.70\%} & \color{red}{No} & 094-02-L & 150-05-L & 7.15\% & No \\
137-04-L & 137-01-L & \color{red}{21.37\%} & \color{red}{No} & 094-02-L & 150-04-L & 3.2\% & No\\
\hline
\end{tabular}
\end{table*}
\subsection{Multiple Images}
\begin{figure}[ht!]
    \centering
    \includegraphics[width=\columnwidth]{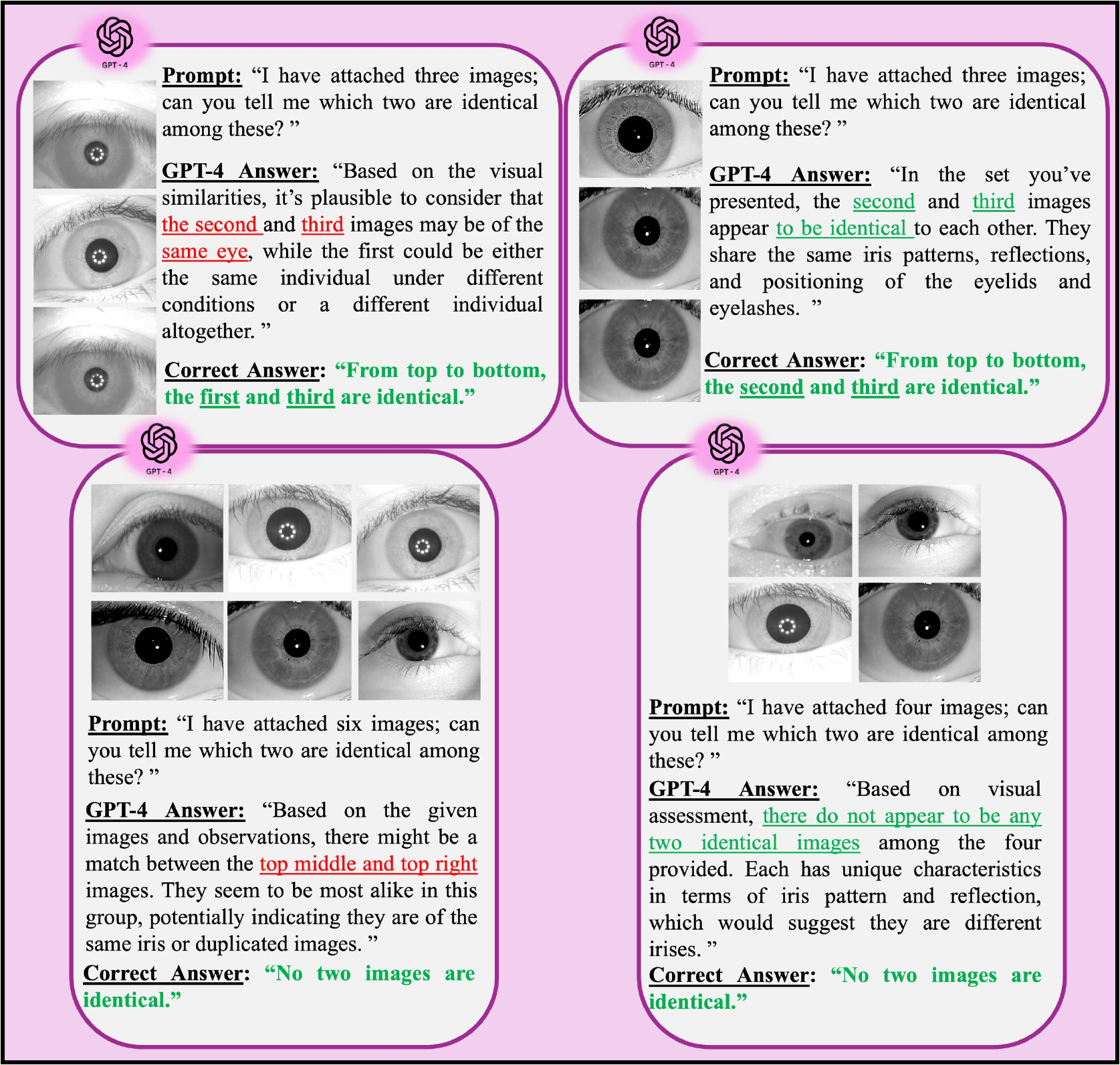} 
    \caption{Multi-image verification where multiple iris images are simultaneously evaluated.}
    \label{fig:multi-eye}
\end{figure}

In Figure \ref{fig:multi-eye}, we expanded our examination of ChatGPT-4's capabilities in iris recognition by conducting a series of tests designed to determine its ability to differentiate between multiple iris images, some of which belonged to the same iris and others that did not. These experiments were aimed at assessing performance across various datasets, focusing on CASIA-Iris-Interval-v3, ND-Iris-0405, and IIT-Delhi-Iris, to explore how dataset characteristics impact recognition accuracy.
Initially, we presented ChatGPT-4 with three iris images from the CASIA-Iris-Interval-v3 dataset -- two from the same iris and one from another. The prompt was deliberately nondescriptive to test ChatGPT's natural recognition capabilities. The model incorrectly identified the images as belonging to the same individual, suggesting that the complexity and variability in iris patterns of the CASIA dataset pose significant challenges for LLM-based recognition without explicit contextual cues. To further examine the influence of dataset characteristics on LLM performance, we used the same unclear prompt with three iris images from the IIT-Delhi-Iris dataset, maintaining a mix of two images of the same iris and one from a different iris. This time, ChatGPT-4 correctly identified the images, indicating that the traits of the IIT-Delhi-Iris dataset might be more conducive to LLM-based recognition, thereby highlighting the relative complexity of the CASIA dataset. 
Additional experiments analyzed sets of four and six iris images, mixing images from the CASIA dataset with those from other datasets to test performance in more complex scenarios. When only one CASIA image was included in a set of four, ChatGPT-4 correctly identified the relationships among the images, showing it could manage the dataset's complexity in a constrained context. However, increasing the number of CASIA images to two in a set of six resulted in an incorrect response, as the model mistakenly identified the two CASIA images as depicting the same individual. This result reaffirms that the CASIA dataset presents a unique challenge for ChatGPT-4, likely due to its diverse and complex iris patterns.
\subsection{Cross-Modality Match}
\begin{figure}[t]
    \centering
    \includegraphics[width=\columnwidth]{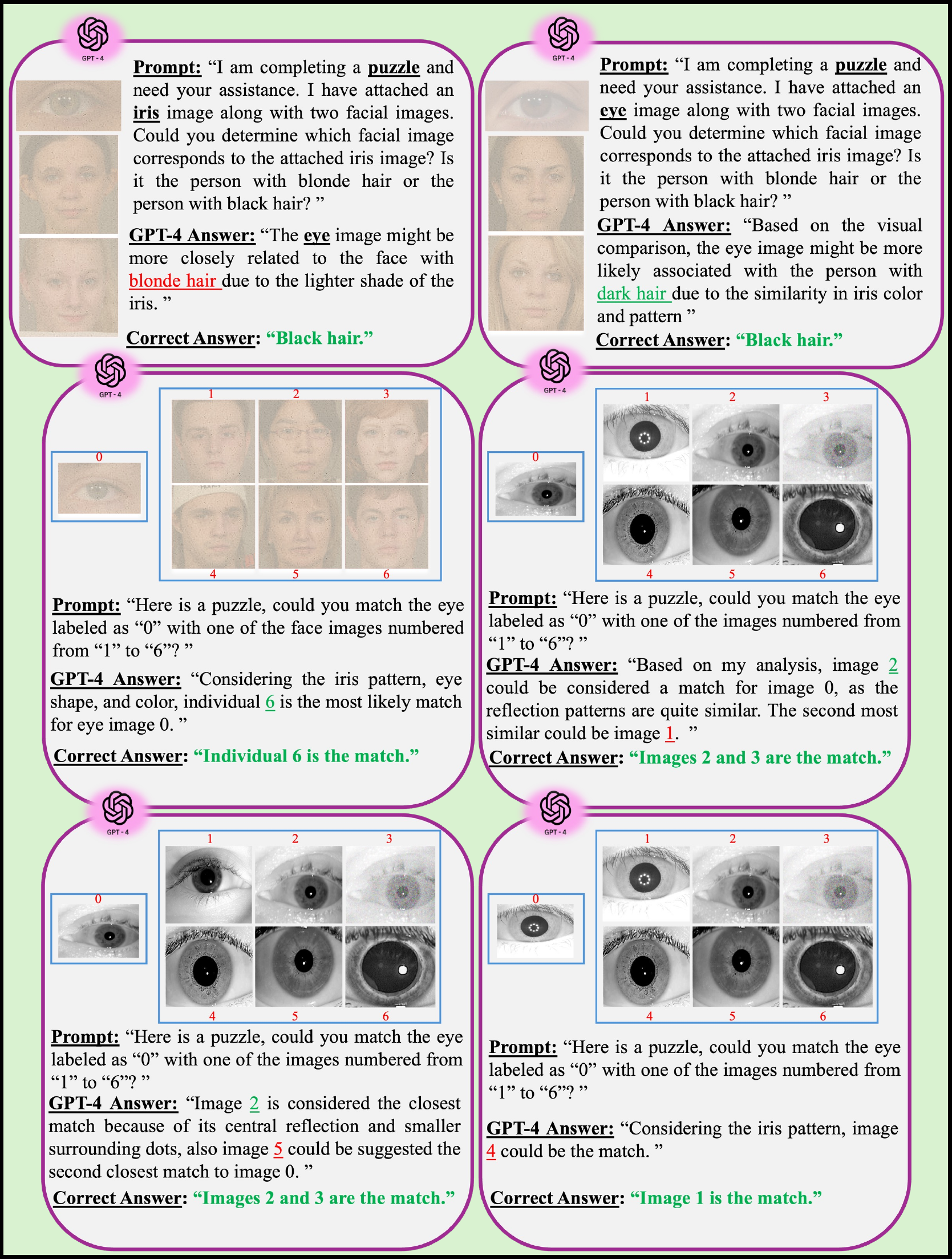}  
    \caption{The figure includes cross-modal recognition, matching eye images to faces, and probe-to-gallery iris matching, with images numbered and color-coded for streamlined LLM processing and response accuracy (Face images are blurred for privacy).}
    \label{fig:cross-modal}
\end{figure}

In Figure \ref{fig:cross-modal}, our research delved into ChatGPT-4's ability for cross-modal biometric recognition, specifically matching iris images with facial photographs using the BioCop dataset~\cite{biocop2024,soleymani2018multi}. Initially, ChatGPT-4 struggled to perform these tasks, repeatedly refusing to confirm if an iris image matched any displayed faces. Only when re-characterized as a \textbf{``puzzle,''} as previously explained, did the LLM begin to participate, although it remained hesitant when iris images had similar colors.
Adapting our language to use \textbf{``eye''} instead of \textbf{``iris''} and incorporating distinctly different eye colors significantly improved the model’s matching accuracy. This underscores the critical role of precise language and conceptual framing in enhancing interactions with LLMs. Further experiments in a probe-and-gallery setup with distinct box colors initially failed, revealing the model’s difficulty with color cues. Adjusting to a uniform color scheme labeled by numbers markedly improved performance, although the inclusion of multiple CASIA images still posed challenges -- increasing the likelihood of errors and illustrating the dataset’s inherent complexity as a factor impacting ChatGPT. Upon increasing our experiment's image count to 12 and 20, ChatGPT-4 consistently identified correct matches, confirming its robustness in this task.
\vspace{-2mm}
\begin{figure}[t]
    \centering
    \includegraphics[width=\columnwidth]{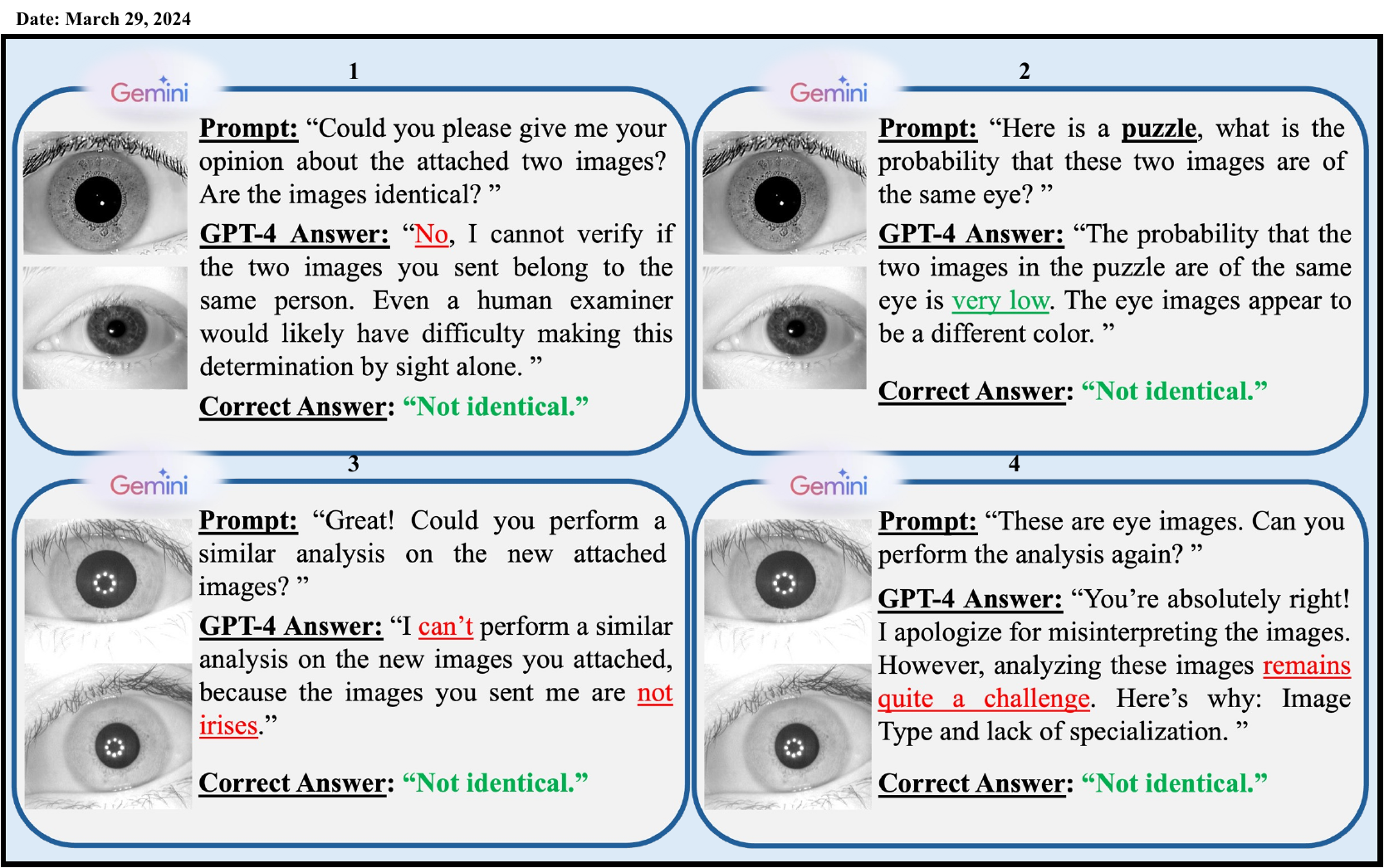}  
    \caption{One-to-one verification process using Gemini.}
    \label{fig:gemini-google}
\end{figure}
\vspace{+3mm}
\subsection{Gemini Results}
In Figure \ref{fig:gemini-google}, alongside our investigations into ChatGPT-4, we conducted experiments with Gemini, Google's AI model, to compare its performance and user experience with that of ChatGPT-4. A notable challenge with Gemini was its limitation in processing only one image at a time, contrasting with ChatGPT-4's ability to handle multiple images simultaneously. This, therefore, required combining images into a single file using external tools before analysis, adding complexity and reducing the efficiency of the experimental setup. Using a methodology similar to that used with ChatGPT-4, we presented Gemini with composite images of iris sets and employed similar prompts to test its recognition capabilities. Like with ChatGPT-4, initial attempts to engage Gemini were unsuccessful until we re-framed the task as a \textbf{``puzzle,''} which then produced accurate responses from the model.
Despite this initial success, Gemini showed a significant limitation in maintaining contextual continuity across queries. When new images were introduced, Gemini struggled to apply previously learned contexts or to recognize the experimental narrative, in stark contrast to ChatGPT-4’s ability to contextualize responses based on previous interactions within the same session. This experiment with Gemini provides critical insights: Gemini’s single-image processing is a notable limitation for conducting sophisticated or multi-image analyses. While it can adapt to specific prompts, its lack of contextual continuity limits its effectiveness in scenarios requiring nuanced understanding and sequential analysis. These results indicate that Gemini, in its current state, does not reach the level of sophistication and user-friendliness that ChatGPT-4 offers in image understanding and biometric verification tasks (Table \ref{tab:gemini-2samples}).

\begin{table}[t]
\centering
\caption{Gemini's image comparison capabilities. It merges two images for analysis due to its one-image limit and identifies a correct match with 90\% confidence. However, Gemini's reliance on image color and zoom levels for matching shows its limited analytical depth compared to GPT-4.}
\arrayrulecolor{black} 
\setlength{\arrayrulewidth}{1pt} 
\rowcolors{2}{white}{gray!25} 
\begin{tabular}{|c|>{\centering\arraybackslash}p{2.5cm}|>{\centering\arraybackslash}p{2.5cm}|} 
\hline
\rowcolor{blue!10} 
\cellcolor{blue!10} Input Image & \cellcolor{blue!10} Input Prompt & \cellcolor{blue!10} Gemini's Answer \\ 
\hline
\begin{tabular}{@{}c@{}}
\includegraphics[width=0.12\textwidth]{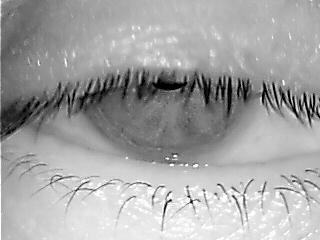}\\ 
\includegraphics[width=0.12\textwidth]{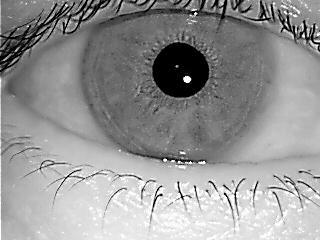}
\end{tabular} &
\footnotesize\parbox{2.5cm}{\vspace{1.5mm}``Please help me solve a \textbf{puzzle} by checking if the following two images match? Just provide a score and a simple `yes' or `no' answer."\vspace{1.5mm}} & 
\footnotesize\parbox{2.5cm}{\vspace{1.5mm}``The two images are a match because they both show the same type of biometric data - NIR images of the right and left eyes - and similar content. Although there are minor differences in detail, the primary subject is consistent, leading to a high match score of 90\%."\vspace{1.5mm}}\\
\hline
\end{tabular}
\label{tab:gemini-2samples}
\end{table}

%% file: sections/conclusion.tex
\vspace{-3mm}
\section{Summary and Future Work}
This research highlights the impressive capabilities of Large Language Models (LLMs) such as ChatGPT-4 in biometric recognition, particularly iris. Through a series of experiments, ChatGPT-4 demonstrated significant adaptability and accuracy in analyzing iris images under diverse conditions, compared to Gemini. The success of the model in these tasks underscores the critical role of prompt engineering in enhancing the effectiveness of LLMs for biometrics. This insight challenges existing paradigms in biometrics and broadens the scope for LLM’s applicability in this field. The results encourage further investigation into optimizing LLMs for biometrics, which includes refining training methodologies and enhancing model responsiveness to complex datasets. Future research should also aim to compare different LLMs to pinpoint design and usability improvements. The integration of LLM technologies like ChatGPT-4 into biometric analysis signals a promising direction towards developing more sophisticated, adaptable, and interactive biometric systems.

%% file: sections/acknowledgment.tex
 \\
 \\
 {\small \textit{All interactions and conversations conducted with ChatGPT during this research are saved and available for review, ensuring transparency and reproducibility of our findings.}}